\documentclass{article}

\usepackage{PRIMEarxiv}

\usepackage[utf8]{inputenc} 
\usepackage[T1]{fontenc}    
\usepackage{hyperref}       
\usepackage{url}            
\usepackage{booktabs}       
\usepackage{amsfonts}       
\usepackage{nicefrac}       
\usepackage{microtype}      
\usepackage{lipsum}
\usepackage{fancyhdr}       
\usepackage{graphicx}       

\graphicspath{{media/}}     

\title{Can Feature Engineering Help Quantum Machine Learning for Malware Detection?}

\author{
  Ran Liu \\
  Univ. of Maryland, Baltimore County \\
  \texttt{\{rliu2@umbc.edu} \\
   \And
  Maksim Eren \\
  Univ. of Maryland, Baltimore County \\
  \texttt{meren1@umbc.edu} \\
   \And
  Charles Nicholas \\
  Univ. of Maryland, Baltimore County \\
  \texttt{nicholas@umbc.edu} \\
}

\begin{document}
\maketitle

\begin{abstract}
With the increasing number and sophistication of malware attacks, malware detection systems based on machine learning (ML) grow in importance. At the same time, many popular ML models used in malware classification are supervised solutions. These supervised classifiers often do not generalize well to novel malware. Therefore, they need to be re-trained frequently to detect new malware specimens, which can be  time-consuming. Our work addresses this problem in a hybrid framework of theoretical Quantum ML, combined with feature selection strategies to reduce the data size and malware classifier training time.
\end{abstract}

\section{Introduction}
With the increasing number and sophistication of malware attacks, malware detection systems based on machine learning (ML) grow in importance. At the same time, many popular ML models used in malware classification are supervised solutions. These supervised classifiers often do not generalize well to novel malware. Therefore, they need to be re-trained frequently to detect new malware specimens, which can be  time-consuming. Our work addresses this problem in a hybrid framework of theoretical Quantum ML, combined with feature selection strategies to reduce the data size and malware classifier training time.

Theoretically, Quantum Parallelism can provide a potential speed-up in a number of ML algorithms. Prior research suggests that an N-dimensional dataset can be mapped onto quantum states with logN qubits. Thus, classical data takes O(logN) steps for the mapping process\cite{Giovannetti2007QuantumMemory}. In addition, quantum operation shows advantages on some specific steps, which include O(poly(log N)) time complexity for Quantum Fourier Transform and matrix inversion, and O(log N) for distance estimation and inner product\cite{Nielsen2012QuantumInformation}. Prior work has also shown the advantages of using quantum algorithms in clustering problems. Harrow et al achieved runtime complexity of O(log(MN)) with a quantum algorithm for clustering, while in comparison, the best classical clustering algorithm takes O(poly(MN))\cite{Harrow2008QuantumEquations}. More specifically, current research on Quantum Annealing shows advantages on some specific ML  algorithms with partial noise robustness ,which limits its application on some practical problems when the dataset is large. In this research, we explored a feature selection strategy based on classical ML methods to help train classifiers on quantum gate model machines (e.g. IBM’s Quantum Processor) when a machine has very limited qubits capacity. We will examine our approach on some state-of-the-art quantum algorithms to be run on gate model machines.

\section{Qubit}\label{sec:Qubit}

‘Qubit’ or ‘Quantum bit’ is the fundamental analogous concept in Quantum computation, which is defined in the Hilbert Space. Using Dirac notation, given an orthonormal basis in Hilbert Space, qubits can be measured to a basis state with probability. We can express a quantum system using state vector language. For example, we can generate superpositions by form linear combinations of states with computational basis $|0\rangle$ and $|1\rangle$:
\begin{equation}\label{eq:1.1}
  |\varphi \rangle = \alpha |0\rangle + \beta |1 \rangle  
\end{equation}
where $\alpha, \beta \in \mathbb{C}$, and $|\alpha|^2 +|\beta|^2 = 1$.
Unlike classical bits which can only be in state 0 or 1, qubit can be in the continuum of basis until it has been observed. For example, \ref{eq:1.1} means qubit can be in state the $|0\rangle$ with probability $|\alpha|^2$  or in the state $|1\rangle$ with probability $|\beta|^2$.
\section{Variational Quantum Classifier}\label{sec:vqc}
In our research, we use the Variational Quantum Classifier proposed by Havlíček, V., Córcoles, A.D., Temme, K. et al. \cite{Havlicek2019SupervisedSpaces}. A classical data point in the interval $\left(0,2\pi\right]$ is first mapped onto the Bloch sphere using the non-linear circuits constructed as:
 \[
\mathcal{U}_{\Phi} (\vec{x}) = \mathit{U}_{\Phi (\vec{x})}\mathit{H}^{\otimes n}\mathit{U}_{\Phi (\vec{x})}\mathit{H}^{\otimes n}
 \]
H is the Hadamard gate and 
\[
\mathit{U}_{\Phi (\vec{x})} = \exp\left( i \sum_{S\subseteq [n] } \phi_S \left(\vec{x}\right) \prod_{\mathit{i} \in S} Z_i\right)
 \]
Then variational circuits are constructed for the optimization purpose, which is called as the \textit{l}-layer circuits. The \textit{l} means repeat constructed variational circuits \textit{l} times. Next the Z-basis measurements  $f:\{0,1\}^n \longrightarrow \{+1, -1\}$are applied to the output bit strings. In the end, measurements are repeated n times to give the probability distribution. And a label as assigned to the input vectors.

\section{Related Work} \label{sec:relatedwork}
\begin{figure}
    \centering
    \includegraphics[scale=0.12]{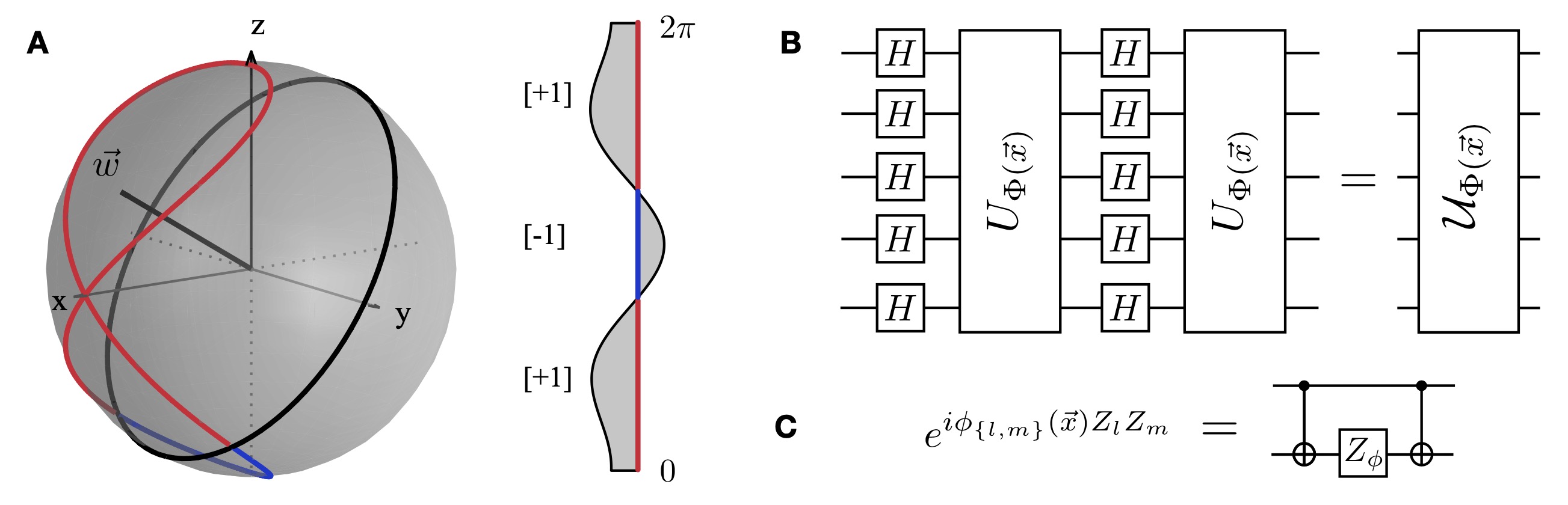}
    \caption{A classical data point in the interval $\left(0,2\pi\right]$ in A is mapped onto the Bloch sphere using the non-linear circuits in B \cite{Havlicek2018}}
    \label{fig:SVMQKenelFunc}
\end{figure}

The Support Vector Machine (SVM) algorithm is a kind of supervised learning algorithm which is used to solve binary classification problems. SVM works on finding a hyperplane to distinguish instances from two different classes. As a hyperplane be defined as:
\[
\vec{w}\cdot \vec{x} + b = 0 
\]
where $\vec{x}$ is the normal vector and $b$ is the offset. Thus, for class A and B, the well trained SVM hyperplanes can give:
\[
\vec{w}\cdot \vec{x} + b \geq 1 if \vec{x} \in A
\]
\[
\vec{w}\cdot \vec{x} + b \leq -1 if \vec{x} \in B 
\]
Thus, the SVM problem is to maximize the margin of two classes A, B s.t. 
\[
y\left(\vec{w}\cdot \vec{x} + b \right) \geq 1
\]
The classification result of a new vector $\vec{x_0}$ is given by:
\[
sign\left(\vec{w}\cdot \vec{x_0} + b \right)
\]
Havlíček, V., Córcoles, A.D., Temme, K. et al. \cite{Havlicek2019SupervisedSpaces} proposed two SVM type classifiers to use the Hilbert Space as the feature space to achieve quantum advantages. The first classifier is to construct the variational circuits that are discussed in \cite{Mitarai2018} \cite{Farhi2018} to generate the hyperplane in the Hilbert Space. The other classifier is to use the quantum computer to directly estimate the kernel function of the Hilbert Space. The quantum mechanism can give quantum advantages if circuits are hard to be simulated by a conventional computer. Their features mapping strategy can be visualized by Figure \ref{fig:SVMQKenelFunc}.
 A classical data point in the interval $\left(0,2\pi\right]$ is first mapped onto the Bloch sphere using the non-linear circuits constructed as:
 \[
\mathcal{U}_{\Phi} (\vec{x}) = \mathit{U}_{\Phi (\vec{x})}\mathit{H}^{\otimes n}\mathit{U}_{\Phi (\vec{x})}\mathit{H}^{\otimes n}
 \]
H is the Hadamard gate and 
\[
\mathit{U}_{\Phi (\vec{x})} = \exp \left( i \sum_{S\subseteq [n]} \phi_S \left(\vec{x}\right) \prod_{\mathit{i} \in S} Z_i\right)
\]

In their first approach, the classical data is first mapped onto the Hilbert Space by the feature mapping strategy described above. Then variational circuits are constructed for the optimization purpose, which they called it as the \textit{l}-layer circuits. The \textit{l} means repeat constructed variational circuits \textit{l} times. Next the Z-basis measurements  $f:\{0,1\}^n \longrightarrow \{+1, -1\}$are applied to the output bit strings. In the end, measurements are repeated n times to give the probability distribution. And a label as assigned to the input vectors. In their second approach, the quantum computer is directly used for the estimated purposes. The quantum computer is first involved in the training process to estimate the kernel of all data pairs in the training dataset. Then after the optimization problem of SVM has been organized, the quantum computer is involved again to estimate the kernel for a new datum of all support vectors. They experiment both approaches using generated customized data. Their results show a 100\% success rate even in the presence of noise.

Patrick Rebentrost, Masoud Mohseni et al. \cite{Rebentrost2014} proposed a quantum approach to analyze the least - squares approximation of SVM. On a high level, they tried to get the optimization parameters by solving the linear equation:
\[
F  \begin{pmatrix}
b\\
\vec{a}
\end{pmatrix}
\equiv \begin{pmatrix}
0 \& 1\\
I \& K + \gamma^{-1}I_N
\end{pmatrix}
\begin{pmatrix}
b\\
\vec{a}
\end{pmatrix}
=
\begin{pmatrix}
0\\
\vec{y}
\end{pmatrix}
\]
where K is the kernel of paris in the training dataset, a is the weight, b is the basis and $\vec{y} = (y_1,\dots, y_M)^T$ are labels. So Matrix F is 
$(M + 1) \times (M + 1)$
Their approach can be summarized as follows: First the classical data points are mapped to the Hilbert space:
\[
|\vec{x_i}\rangle = \frac{1}{|\vec{x_i}|} \sum_{i = 1}^M (\vec{x_i})_j|j\rangle 
\]
Second, the HHL algorithm \cite{Aram2009} is applied to get the optimized hyperplane parameter $\vec{a}$ and $b$. Then the quantum register is transformed to:
\[
|b, \vec{a}\rangle = \frac{1}{\sqrt{C}} (b|0\rangle + \sum_{k = 1}^M (a_k|k\rangle )
\]
where $C = b^2 + \sum_{k = 1}^M a_k^2 $.
The classification result is given by:
\[
y(\vec{x_0}) = (\sum_{i = 1}^M a_i (\vec{x_i} \cdot \vec{x_0}) + b)
\]
\section{Experiment}
In our experiments, we use the Drebin dataset consisting of +15k benign and malware samples with 215 dynamic and static analysis based features\cite{Yerima2019DroidFusion:Detection}. We utilize XGBoost and Decision Tree (DT) to identify the top 20 most important features. Using these features, we next conducted our experiments on the IBM Qiskit Simulator with Variational Quantum Classifier (VQC). The preliminary results show that VQC with XGBoost selected features can get a 78.91\% test accuracy score on 10,000 samples with a 50\% test set split. Differently, VQC with DT feature selection got 62.41\% test accuracy. We also conducted tests using IBM 5 qubits machines. Our experiments in this round were limited by the number of qubits in IBM Quantum Machine. Therefore we ran our experiment ten times with 20 randomly selected samples on each run with a 50\% test set split. To show that our results are statistically significant, we report our final accuracy score with a 95\% Confidence Interval (CI), using QSVM. The final average accuracy for the model trained using the features selected with XGBoost was 74\% (+- 11.35\%).

We next explored oversampling and under-sampling strategies based on classical ML methods to help train classifiers on quantum gate model machines for malware detection. Our preliminary experimental results show:
\begin{itemize}
    \item In the case of using oversampling techniques, the accuracy increased up to 80.15\% from 74\% with SMOTE,  and up to 78.06\% with Adaptive Synthetic (ADASYN).
    \item In the case of using under-sampling techniques, the accuracy increased up to 78.62\% from 74\% by removing samples using K-means, and up to 77.26\% by removing samples which do not agree “enough” with their neighborhood.
    \item To overcome noisy samples introduced by SMOTE, we combined the use of oversampling and under-sampling, the accuracy increased up to 83.78\%.
    \item We conducted our experiments on the IBM 5 qubits Quantum Machine. Due to the limited number of qubits available on the Quantum Machine, we use 20 samples with a 50\% test set split. The Quantum SVM got 56\% accuracy and VQC achieved 80\% success for malware detection. 
\end{itemize}

\section{Conclusion}
In this study, we explored the use of feature engineering and the quantum machine for malware detection tasks. The practical use of a quantum machine for malware detection is still in its early stages due to the physical limitations.We hope that our work will inspire other researchers to explore this direction.

\bibliographystyle{unsrt}
\bibliography{local, Mendeley}

\end{document}